\documentclass[final,1p,times]{elsarticle}

\usepackage{amssymb}

\usepackage{amsmath,amsfonts,amsthm}
\usepackage{algorithm}
\usepackage{algorithmic}
\usepackage{bm}
\usepackage{bbm}
\usepackage{float} 
\usepackage[symbol]{footmisc}
\usepackage{graphicx}
\usepackage[utf8]{inputenc}
\usepackage{makecell}
\usepackage{multirow}
\usepackage{url}
\usepackage{xcolor}
\usepackage{caption}
\usepackage{subcaption}
\newcommand{\rulesep}{\unskip\ \vrule\ }

\usepackage{lineno}
\modulolinenumbers[5]

\definecolor{red}{rgb}{.9,0.1,0.1}

\journal{Applied Soft Computing}

\begin{document}

\begin{frontmatter}



\title{Towards Robust Partially Supervised Multi-Structure Medical Image Segmentation on Small-Scale Data}

\author[1]{Nanqing Dong\corref{cor1}}
\cortext[cor1]{Corresponding author}
\ead{nanqing.dong@cs.ox.ac.uk}
\author[2]{Michael Kampffmeyer}
\author[3]{Xiaodan Liang}
\author[4]{Min Xu}
\author[1]{Irina Voiculescu}
\author[5,6]{Eric Xing}

\affiliation[1]{organization={Department of Computer Science, University of Oxford},
            city={Oxford, England},
            country={UK}}
\affiliation[2]{organization={Machine Learning Group, UiT The Arctic University of Norway},
            city={Troms{\o}},
            country={Norway}}
\affiliation[3]{organization={School of Intelligent Systems Engineering, Sun Yat-sen University},
            city={Guangzhou, Guangdong},
            country={China}}
\affiliation[4]{organization={Computational Biology Department, Carnegie Mellon University},
            city={Pittsburgh, Pennsylvania},
            country={USA}}
\affiliation[5]{organization={Machine Learning Department, Carnegie Mellon University},
            city={Pittsburgh, Pennsylvania},
            country={USA}}
\affiliation[6]{organization={Mohamed bin Zayed University of Artificial Intelligence},
            city={Masdar City, Abu Dhabi},
            country={UAE}}

\begin{abstract}
The data-driven nature of deep learning (DL) models for semantic segmentation requires a large number of pixel-level annotations. However, large-scale and fully labeled medical datasets are often unavailable for practical tasks. Recently, partially supervised methods have been proposed to utilize images with incomplete labels in the medical domain. 
To bridge the methodological gaps in partially supervised learning (PSL) under data scarcity, we propose \textit{Vicinal Labels Under Uncertainty} (VLUU), a simple yet efficient framework utilizing the human structure similarity for partially supervised medical image segmentation. Motivated by multi-task learning and vicinal risk minimization, VLUU transforms the partially supervised problem into a fully supervised problem by generating vicinal labels. We systematically evaluate VLUU under the challenges of small-scale data, dataset shift, and class imbalance on two commonly used segmentation datasets for the tasks of chest organ segmentation and optic disc-and-cup segmentation. The experimental results show that VLUU can consistently outperform previous partially supervised models in these settings. Our research suggests a new research direction in label-efficient deep learning with partial supervision.
\end{abstract}




\end{frontmatter}


\section{Introduction}
\label{sec:intro}
Convolutional Neural Networks (CNNs) have been a game-changer for the task of semantic segmentation \cite{long2015fully,ronneberger2015u,chen2017deeplab}, as they can learn pixel-level mappings from the image space to the label space via end-to-end training. To learn these complex mappings, state-of-the-art CNNs usually leverage a large number of parameters and require the availability of large-scale fully labeled datasets, which are often unavailable for real-life tasks. In the medical domain, where annotations require substantial efforts from clinical experts, obtaining these datasets can be challenging. 
This has led to an increasing interest in learning from partially labeled data, when fully labeled data is not available. \textit{Partially supervised learning} (PSL) is still an open research question in medical image segmentation \cite{gonzalez2018multi,petit2018handling,zhou2019prior,fang2020multi,shi2021marginal}. From the perspective of multi-task learning (MTL) \cite{caruana1997multitask}, a semantic segmentation task can be decomposed into multiple sub-tasks corresponding to each semantic class of interest, which provides the theoretical foundations of learning from partial ground truth.
Given a medical image segmentation task with multiple classes of interest, it is common to collect and merge several \textit{available}, smaller but relevant datasets into a larger dataset under the challenges of small-scale data, dataset shift, and class imbalance. These smaller datasets were originally labeled for sub-tasks, such that only the objects related to the specific sub-task are annotated, while other objects are merged into the background. In other words, the training images do not have complete annotations for all classes of interest but are \textit{partially labeled}. For example, in the task of abdominal organ segmentation, a pancreas dataset and a liver dataset might be available separately, where only the pancreas and the liver are labeled, respectively.

A key challenge, leading to poor segmentation performance when considering multiple partially labeled datasets, is that the semantic classes of one dataset could be categorized as the background for another dataset that was annotated for a different purpose. Traditional semantic segmentation models \cite{long2015fully,ronneberger2015u,chen2017deeplab} can therefore not be directly applied and trained end-to-end in a supervised fashion. Further, given the small amount of partially labeled data, deep learning (DL) models are prone to overfitting. 

Recent studies in PSL~\cite{gonzalez2018multi,petit2018handling,zhou2019prior,dmitriev2019learning,fang2020multi,shi2021marginal} all assume that, for each class of interest, enough training examples are accessible. However, considering the data scarcity in most practical medical tasks, usually, only few training examples might be available, making previous approaches impractical.

\begin{figure*}[th]
    \centering
    \begin{subfigure}[t]{0.15\textwidth}
        \centering
        \includegraphics[width=\textwidth]{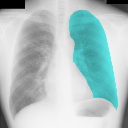}
        \caption{}
    \end{subfigure}
    \begin{subfigure}[t]{0.15\textwidth}
        \centering
        \includegraphics[width=\textwidth]{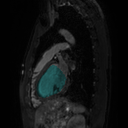}
        \caption{}
    \end{subfigure}
    \begin{subfigure}[t]{0.15\textwidth}
        \centering
        \includegraphics[width=\textwidth]{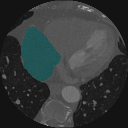}
        \caption{}
    \end{subfigure}
    \begin{subfigure}[t]{0.15\textwidth}
        \centering
        \includegraphics[width=\textwidth]{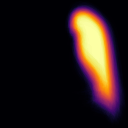}
        \caption{}
    \end{subfigure}
    \begin{subfigure}[t]{0.15\textwidth}
        \centering
        \includegraphics[width=\textwidth]{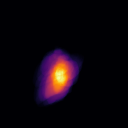}
        \caption{}
    \end{subfigure}
    \begin{subfigure}[t]{0.15\textwidth}
        \centering
        \includegraphics[width=\textwidth]{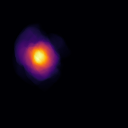}
        \caption{}
    \end{subfigure}
    \caption{Annotated examples of different type of medical images (first row): (a) a posteroanterior X-ray image with the ground truth annotation of the left lung; (b) a sagittal MRI image with the ground truth annotation of the left ventricle; (c) an axial CT image with the ground truth annotation of the right atrium. The label distributions (normalized density heatmap) of the corresponding organs in public datasets (second row): (d) the left lungs in the JSRT dataset~\cite{jsrt}; (e) the left ventricles in the MRI-WHS dataset~\cite{zhuang2010registration}; (f) the right atriums in the CT-WHS dataset~\cite{zhuang2015multiatlas}.}
    \label{fig:1}
\end{figure*}

To bridge the methodological gaps when only small-scale partially labeled data is available, we propose a simple yet efficient framework \textit{Vicinal Labels Under Uncertainty} (VLUU) by exploring the statistical similarity of human structures (e.g.~shape, size, location) among different patients. See Fig.~\ref{fig:1} for an illustration of such a similarity. The proposed framework is motivated by vicinal risk minimization (VRM) \cite{chapelle2001vicinal}, where the fully labeled vicinal examples are generated by linearly combining randomly sampled partial labels with a weight randomly sampled from a Dirichlet distribution. These vicinal examples allow us to transform the partially supervised problem into a fully supervised one. That is to say, we can utilize any existing supervised segmentation networks and loss functions to solve partially supervised problems. The generated vicinal labels contain uncertainty regions where classes of interest could potentially overlap. We utilize these uncertainties in the training process to improve the robustness of DL models.  

Recent studies have shown that VRM can consistently improve the performance of CNNs for image classification tasks~\cite{zhang2018mixup,yun2019cutmix}. However, there is a lack of definition of VRM for dense prediction tasks with incomplete labels, e.g.~\cite{zhang2018mixup} and~\cite{yun2019cutmix} can not be directly applied on partially supervised semantic segmentation tasks. Instead, we revisit VRM, a long-ignored but particularly efficient approach, to tackle this problem. Specifically, by defining a generic vicinity distribution, VLUU learns a mapping from a sequence of images to a vicinal label which is generated by statistically \textit{mixing up} the corresponding partial labels of the input images. 

We perform the first systematic study of partially supervised methods under data scarcity challenges, such as small-scale data, \textit{domain shift} or \textit{dataset shift} \cite{quionero2009dataset}, and class imbalance, on two representative medical image segmentation tasks, namely chest organ segmentation and optic disc-and-cup segmentation. The experiments show that VLUU is more robust than previous partially supervised methods under these settings. The proposed framework has five advantages over previous methods: (1) it is easy to implement without relying on complex loss functions, network architectures, and optimization procedures; (2) it can be trained end-to-end in supervised settings with common segmentation networks and loss functions; (3) it does not require any fully labeled images in the training data; (4) it can efficiently reduce the risk of overfitting for small-scale data; and (5) it can be easily extended to adversarial training.

Our main contributions can be summarized as follows:
\begin{enumerate}
\item We propose a simple yet robust framework for partially supervised medical image segmentation, which is robust when there is only limited partially labeled data.
\item We provide theoretical interpretations for the proposed framework based on vicinal risk minimization and multi-task learning.
\item We systematically evaluate the robustness of partially supervised methods and show that the proposed framework can outperform state-of-the-art partially supervised methods under various data scarcity challenges.
\end{enumerate}

The rest of this paper is organized as follows. Sec.~\ref{sec:related} reviews the relevant literature. Sec.~\ref{sec:vrm} and Sec.~\ref{sec:ana} describe the proposed framework and its properties. Sec.~\ref{sec:exp} describes the proposed benchmark task and provides experimental results and analysis. Sec.~\ref{sec:con} summarizes this work.

\section{Related Works}
\label{sec:related}

\subsection{Semi-Supervised Learning}
\label{sec:related_ssl}
In machine learning, semi-supervised learning (SSL) falls between supervised learning (SL), where only fully labeled training data are available, and unsupervised learning (UL), where no labels are available. In semi-supervised learning, the training set consists of both labeled and unlabeled data. The robust state-of-the-art semi-supervised methods include label propagation (LP) \cite{iscen2019label}, graph neural networks \cite{kipf2017semi,jiang2019semi}, and cross consistency training \cite{ouali2020semi}. Most semi-supervised methods can not be applied to PSL problems directly as they are required to minimize a supervised loss, however, among these seminal SSL methods, LP \cite{wang2007label} can be applied to tackle partially labeled data directly. With LP, pseudo-labels are generated based on prior information (partially labeled data). Then, the pseudo-labels are fine-tuned iteratively toward convergence \cite{zhu2002learning}. LP is computationally expensive and the quality of the pseudo-labels is highly dependent on the number of training data. \cite{zhou2019prior} has demonstrated that LP is a powerful solution to PSL with fully labeled datasets as prior. As a robust method tested by time, LP is a strong baseline in this work.

\subsection{Partially Supervised Learning}
\label{sec:related_psl}
Closely related to SSL, partially supervised learning (PSL), or the partial labels problem, describes the situation where each example has an incomplete label (e.g.~only one semantic class is annotated out of a few classes of interest). Concretely, given a collection of multiple small partially labeled datasets, each dataset may only contain annotations for a \textit{proper subset} of classes of interest and these subsets are disjoint. In such a case, the images in the collection are partially labeled. A more rigorous formulation of the problem is presented in Sec.~\ref{sec:vrm_def}.

PSL is a topic of active research as the perfect fully labeled training datasets tend to be only available for specific research tasks. In recent studies, several methods have been proposed to address semantic segmentation with partial labels from different aspects. \cite{triggs2008scene} treats a grid of image patches as nodes and uses conditional random fields to propagate information. However, as a result, the predicted segmentation masks will be unnatural due to the patch-wise prediction. In DL, a common approach is to treat the missing labels as the background. This approach can be viewed as a naive form of \textit{noisy labels} \cite{natarajan2013learning} and only works when the pixels of missing classes take up a much smaller portion of the images, compared with the pixels of the background. For benchmark datasets in computer vision such as PASCAL VOC \cite{everingham2010pascal} and MS COCO \cite{lin2014microsoft}, there are only a few classes present in each image or the objects can be very small. Thus, merging unlabeled pixels into the background might be an efficient solution for these datasets. In contrast, for common medical datasets, multiple classes can be present in each image and the objects of interest (e.g.~organs) may take up the majority of the pixels. Another common approach in DL is to ignore the cross entropy of the missing classes during backpropagation \cite{gonzalez2018multi,petit2018handling}. The limitation of this approach is that abandoning the pixel information of missing classes means that the learners (CNNs) will receive much less supervision during the learning process, both from the image space and the label space. A direct result is that the learner can not discriminate the classes of interest against the background. Recently, PaNN \cite{zhou2019prior} proposes a complex Expectation-Maximization (EM) algorithm with a primal-dual optimization procedure. However, PaNN requires the availability of fully labeled images as prior, which is often unavailable. To address general semantic segmentation \cite{everingham2010pascal,cordts2016cityscapes}, \cite{dmitriev2019learning} proposes to use a complex encoder-decoder architecture to condition the partial information within the CNN, which requires a large dataset to comply with the large number of parameters. PIPO-FAN \cite{fang2020multi} proposes a complex pyramid feature fusion mechanism and a target adaptive loss (TAL). Unlike the other methods, PIPO-FAN has a demanding requirement in the training process, i.e.~the examples with the same partial labels must be trained together. It is worth mentioning that TAL also treats the missing labels as the background. Recently, a state-of-the-art work \cite{shi2021marginal} tackles PSL by proposing a marginal loss and an exclusion loss, which are designed for partially supervised medical image segmentation. From the perspective of DL, \cite{shi2021marginal} tries to address PSL at the last step of feed-forward propagation, while this work addresses PSL at the data preparation step, which is before the feed-forward propagation process. To sum up, all of these methods are only applicable when substantial partially labeled images or fully labeled images are available. In addition, previous studies do not consider the practical situations such as dataset shift and class imbalance. A detailed empirical analysis is provided in Sec.~\ref{sec:exp_cos}.

\subsection{Multi-Task Learning}
\label{sec:related_mtl}
\begin{figure*}[t]
    \centering
    \begin{subfigure}[t]{0.24\textwidth}
        \centering
        \includegraphics[width=\textwidth]{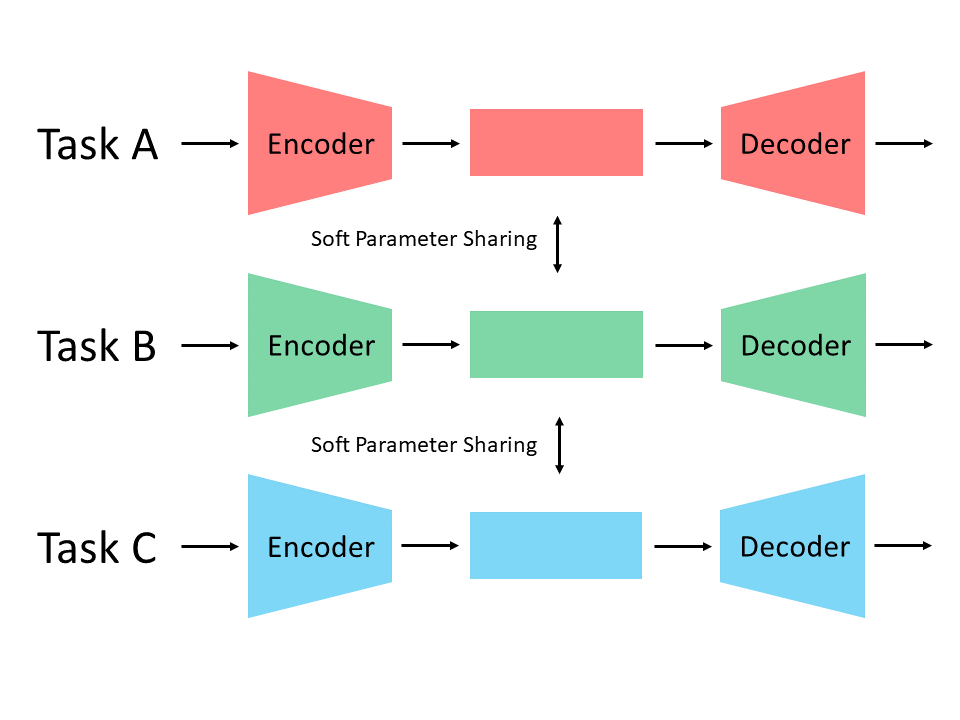}
        \caption{}
    \end{subfigure}
    \begin{subfigure}[t]{0.24\textwidth}
        \centering
        \includegraphics[width=\textwidth]{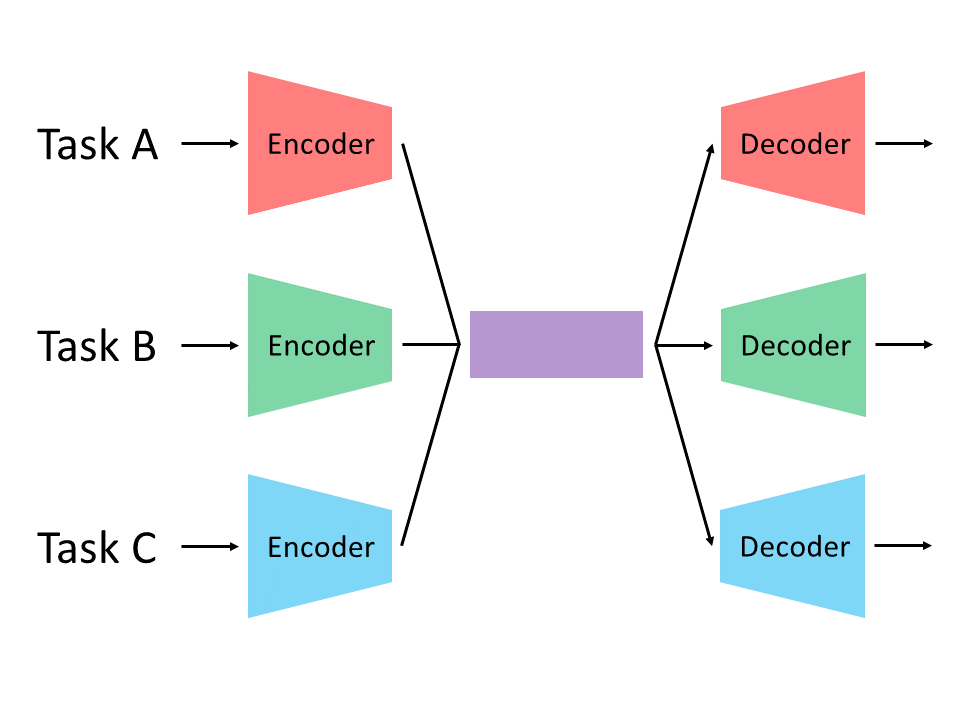}
        \caption{}
    \end{subfigure}
    \begin{subfigure}[t]{0.24\textwidth}
        \centering
        \includegraphics[width=\textwidth]{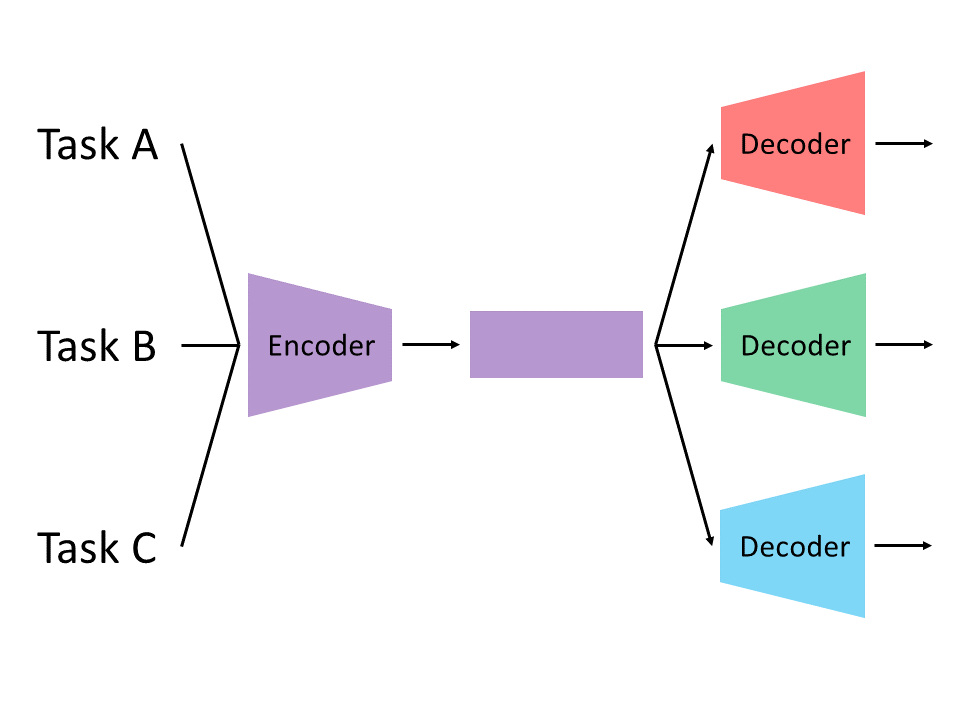}
        \caption{}
    \end{subfigure}
    \begin{subfigure}[t]{0.24\textwidth}
        \centering
        \includegraphics[width=\textwidth]{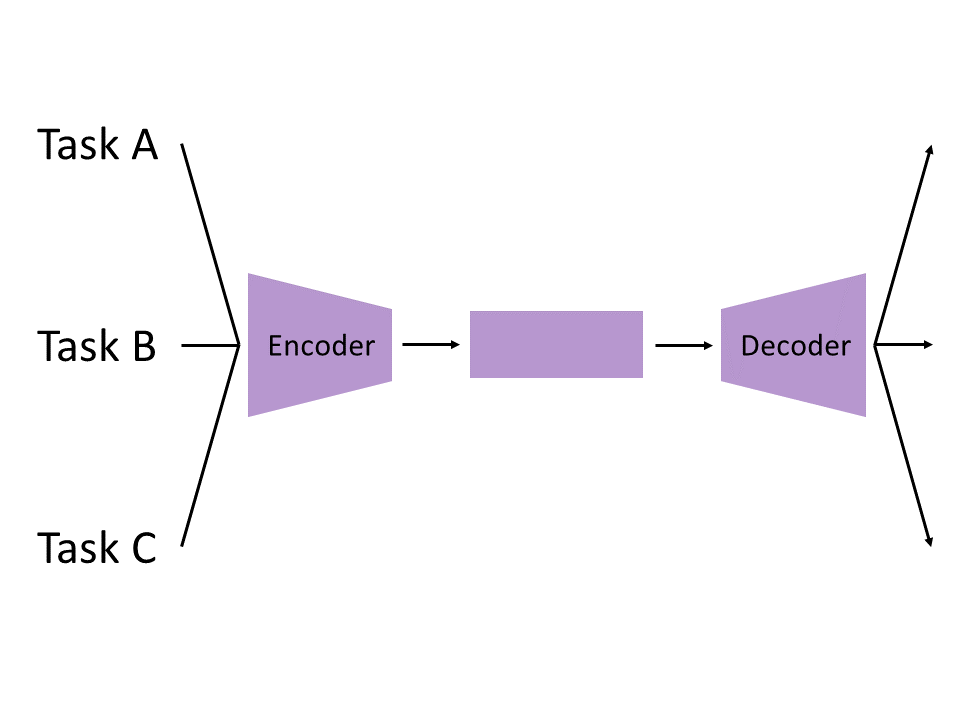}
        \caption{}
    \end{subfigure}
    \caption{Common MTL workflows for dense prediction tasks. The data modalities of the input are different: (a) The different tasks have separate networks, which are linked through \textit{soft parameter sharing}. Note, without soft parameter sharing, (a) depicts the standard multiple single-task learning. (b) The different tasks have independent encoders and decoders but share the same network backbone (in purple), which is also known as \textit{hard parameter sharing}. The data modalities of the input are identical: (c) Each task has independent output, which requires an independent decoder. (d) The tasks can share the same decoder.}
    \label{fig:mtl}
\end{figure*}

By leveraging task-specific information, multi-task learning (MTL) \cite{caruana1997multitask} can improve the model generalization when the tasks of interest are somewhat related. In the era of DL, we aim to use a neural network (NN) to map the input to the output, given a task. In contrast to single-task learning, where each task is handled by an independent NN, MTL can reduce the memory footprint, increase overall inference speed, and improve the model performance. When the associated tasks contain complementary information, MTL can regularize each single task. For dense prediction tasks, a good example is semantic segmentation, where we always assume that the classes of interest are mutually exclusive. Depending on the data modality, task affinity \cite{vandenhende2020branched} between sub-tasks and task fusion strategy, there are various types of MTL. We depict several common MTL workflows that are related to our work in Fig.~\ref{fig:mtl}. Semantic segmentation falls into the category Fig~\ref{fig:mtl}(d). As pointed out by \cite{zhang2019pattern}, pixel-level tasks in visual understanding often have similar characteristics, which can be potentially used to boost the performance by MTL. We argue that PSL problems can be reformulated as MTL problems by utilizing human structure similarity.

\section{Method}
\label{sec:vrm}

\subsection{Preliminaries}
In SL, given a training dataset $S = \{X, Y\}$ with images $X = \{x_i\}_{i=1}^n$ and ground truth labels $Y = \{y_i\}_{i=1}^n$, the empirical risk is defined as 
\begin{equation}
\mathcal{R}(h) = \frac{1}{n}\sum_{i}^{n} L(h(x_i), y_i),
\label{eq:emp_risk}
\end{equation}
where $L(\cdot, \cdot)$ is the loss function and $h \in \mathcal{H}$ is the hypothesis. In this work, we assume that $L$ and $h$ are universal as they can be any loss function and model in a standard supervised setting. For example, for a popular choice of semantic segmentation, $L$ could be the cross entropy and $h$ could be a CNN. The minimization of the empirical risk $\mathcal{R}(h)$ is also known as Empirical Risk Minimization (ERM) in statistical learning literature \cite{vapnik1998statistical}. 

\subsection{Problem Formulation}
\label{sec:vrm_def}
Assume there are $K > 1$ mutually exclusive semantic classes of interest present in the same image, i.e. there is no hierarchical relationship between classes and all classes are present. In this work, we focus on the challenging situation that each image is annotated for only one semantic class. For partially labeled images, we can always split $S$ into $K$ sub-datasets where each sub-dataset contains label information of only one class. Here, the $K$ datasets are mutually exclusive in terms of both images and classes. Mathematically, we have $S = \bigcup_{j}^K S_j$, where $S_j = \{X_j, Y_j\}$ denotes the partially labeled dataset with label information of semantic class $j$. In $S_j$, $X_j = \{x^{j}_{i}\}_{i=1}^{n_j}$ is the image set of the images with label information of the semantic class $j$ and $Y_j = \{y^{j}_{i}\}_{i=1}^{n_j}$ contains the corresponding partial labels. In addition, we define $S_j \subset \mathcal{D}_j$, where $\mathcal{D}_j$ denotes the source domain for $S_j$, and we define $d(\mathcal{D}_{j_1}, \mathcal{D}_{j_2}) \neq 0 \ \forall j_1 \neq j_2$, where $d(\cdot, \cdot)$ measures the distributional discrepancy between two distribution. That is to say, dataset shift exists. As a comparison, previous studies usually fail to validate this assumption when using one fully labeled dataset to simulate the partially labeled datasets.

Note, the problem formulation here describes the most general case as all other cases are trivial extensions. For example, when an image has annotations for more than one semantic class, duplicate image copies could exist in multiple datasets and the above mathematical formulation still holds.

\begin{figure}[t]
    \centering
    \includegraphics[width=0.8\textwidth]{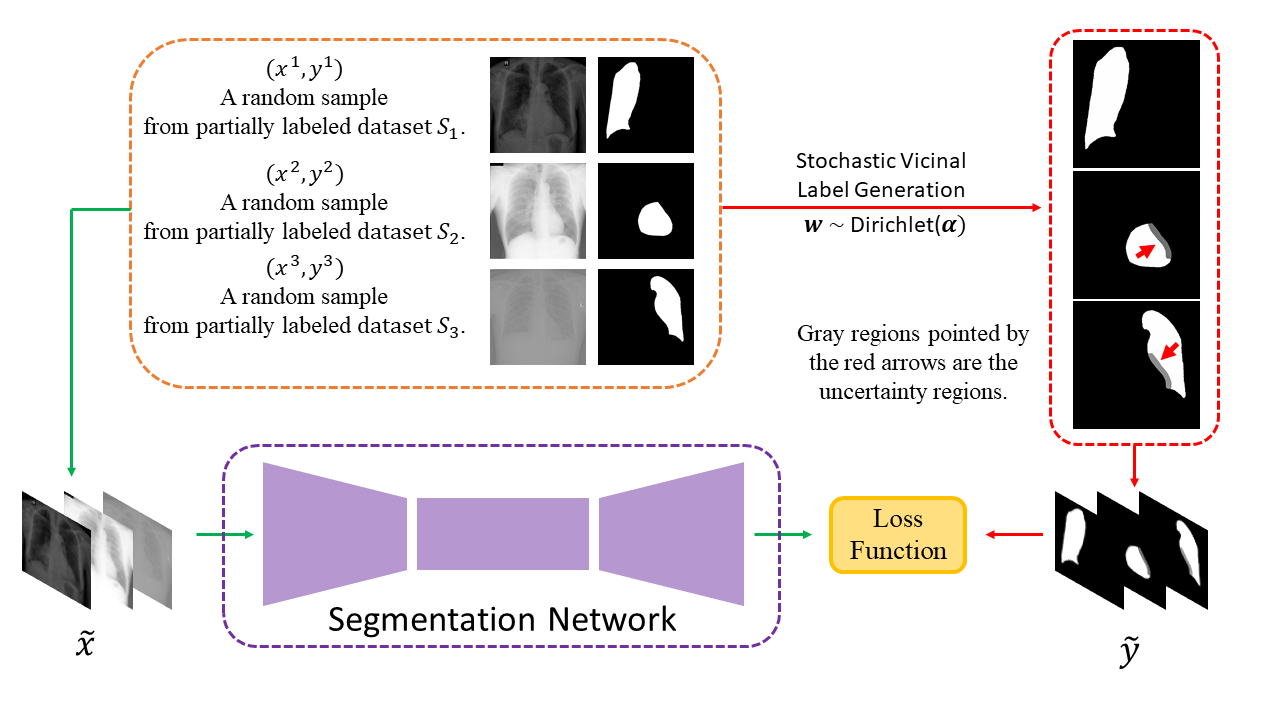}
    \caption{Illustration of the standard training pipeline. Here, we use the chest organ segmentation task as an example. Assume there are three classes of interest, which are left lung, heart, and right lung. And there are three corresponding partially labeled sub-datasets, denoted as $S_1$, $S_2$ and $S_3$. $\{(x^1, y^1), (x^2, y^2), (x^3, y^3)\}$ are randomly sampled from $S_1$, $S_2$ and $S_3$, respectively. The vicinal example pair $(\tilde{x}, \tilde{y})$ is generated by Eq.~\ref{eq:vic_x} and Eq.~\ref{eq:vic_y} with $K=3$. The segmentation network could be any standard segmentation network such as FCN \cite{long2015fully} or U-Net \cite{ronneberger2015u}. For simplicity, the background mask is not shown in the figure and we use grayscale images to visualize the vicinal labels. }
    \label{fig:train}
\end{figure}

\subsection{Vicinal Labels Under Uncertainty}
\label{sec:vrm_vrm}
In a fully supervised setting, introducing statistical randomness \cite{chapelle2001vicinal} and using the convex combination of the training data \cite{zhang2018mixup,yun2019cutmix} are two efficient methods to improve the robustness of DL models. However, as none of these methods can address the missing class information, they have been ignored in multi-class semantic segmentation with partial supervision for a long time. In this work, we integrate and extend these two simple ideas. Instead of designing complex networks \cite{dmitriev2019learning,fang2020multi} or loss functions \cite{shi2021marginal}, we utilize the partial labels in a multi-task fashion. A naive solution is to decompose the partially supervised multi-class segmentation task into multiple binary segmentation tasks. As both the input and the output share the same characteristics, we want to use a shared encoder and decoder, similar to Fig.~\ref{fig:mtl}(d). However, unlike semantic segmentation, where there is only a single image as input and the corresponding label is based on the same image, we now have images and labels from different partially labeled datasets. We propose to \textit{fuse} the tasks based on the human structure similarity. 

Let $x$ be a 2D medical image with size $H \times W$, represented by a 2D array, which has been pre-processed via instance normalization and optional spatial alignment. So $y$ is the corresponding partial label with one semantic class annotated, represented by a 3D array ($H \times W \times (K+1)$), where the last dimension corresponds to the semantic classes. For each pixel in $x$, the corresponding element in $y$ is a $(K+1)$-element one-hot vector for the background and $K$ semantic classes. For simplicity, we use $y[k]$ to denote the binary label map for class $k \le K$ ($k = 0$ denotes the background), which is the $(k+1)^{th}$ semantic channel of $y$. Let $(x^{j}, y^{j})$ be a random sample from $S_j$, and so $\{(x^{j}, y^{j})\}_{j=1}^K$ is a $K$-element tuple of such samples. We define
\begin{equation}
\tilde{x} = concat(\{x^{j}\}_{j=1}^K)
\label{eq:vic_x}
\end{equation}
\begin{equation}
  \tilde{y} =
    \begin{cases}
      \frac{w_k y^{k}[k]}{\sum_{j=1}^K w_j y^{j}[j] + \epsilon} & k > 0\\
      1 - \sum_{j=1}^K \tilde{y}[j] & k = 0
    \end{cases}   
\label{eq:vic_y}  
\end{equation}
, where $concat$ is the \textit{concatenate} operation that concatenate $\{x^{j}\}_{j=1}^K$ along a new dimension. We have $\bm{w} = (w_1, ..., w_K)\sim\text{Dirichlet}(\bm{\alpha})$ with $\bm{\alpha} = (\alpha_1, ..., \alpha_K) \in (0, \infty)^K$ and $\epsilon > 0$ is a small number to ensure numeric stability, e.g.~$\epsilon = 10^{-3}$. Without prior information over the true label distributions, we setup $\bm{\alpha}$ as a constant vector, i.e.~$\alpha_k = \alpha \ \forall 1 \leq k \leq K$. Given $(\tilde{x}, \tilde{y})$, we transform a partially supervised problem into a fully supervised one and we can utilize any existing supervised segmentation network and loss function. See Fig.~\ref{fig:train} for the illustration of the training pipeline. 
In each class channel of the vicinal label, the continuous probabilities are transformed into grayscale pixels for visualization. There are two origins of uncertainty for generating the vicinal labels when there is an overlap between partial labels. First, the sampling of input images is stochastic. Second, $\bm{w}$ is randomly sampled from a Dirichlet distribution (e.g.~$\bm{w} = (0.33, 0.41, 0.26)$ used in Fig.~\ref{fig:train}). See the upper right corner in Fig.~\ref{fig:train} for visual examples intuitively, where $y_2$ and $y_3$ have an overlapping region.

\subsubsection{Theoretical Interpretation}
The proposed solution can be interpreted from two aspects, namely vicinal risk minimization (VRM) \cite{chapelle2001vicinal} and MTL respectively. In VRM, a vicinity distribution $\mathcal{V}$ is defined as the probability distribution for the virtual image-label pair (also known as vicinal example) $(\tilde{x}, \tilde{y})$ in the vicinity of $(x, y)$. The vicinal risk is defined as 
\begin{equation}
\mathcal{R}_{\mathcal{V}}(h) = \frac{1}{n}\sum_{i}^{n} L(h(\tilde{x_i}), \tilde{y_i}).
\label{eq:vic_risk}
\end{equation}
Eq.~\ref{eq:vic_y} factually defines a non-parametric anatomical prior for the label distribution. In state-of-the-art VRM works for image classification \cite{zhang2018mixup,yun2019cutmix}, the vicinal image is usually defined as the convex combination of real images, where the parameters for the convex combination are sampled from statistical distributions. As a comparison, we utilize a CNN ($h$ in Eq.~\ref{eq:vic_risk}) to learn this parametric convex combination jointly with semantic segmentation. Eq.~\ref{eq:vic_x} and the CNN jointly play the role of $\tilde{x}$ in Eq.~\ref{eq:vic_risk}. By combining Eq.~\ref{eq:vic_x} and Eq.~\ref{eq:vic_y}, we inexplicitly define a generic $\mathcal{V}$.

On the other hand, given $K$ sub-tasks, we are using a CNN to learn a $K \mapsto K$ task mapping. Eq.~\ref{eq:vic_y} is a task-fusion process that fuses different but related task knowledge. We want to maximally share the network architecture from a MTL perspective. To achieve this, the novelty here is that we utilize the human structure similarity to \textit{mix up} the partial labels. Meanwhile, the uncertainty regions in the vicinal labels, caused by the stochastic convex combination of partial labels, can reduce the risk of overfitting and improve the robustness when the training data is small.

\begin{figure*}[th]
    \centering
    \includegraphics[width=0.8\textwidth]{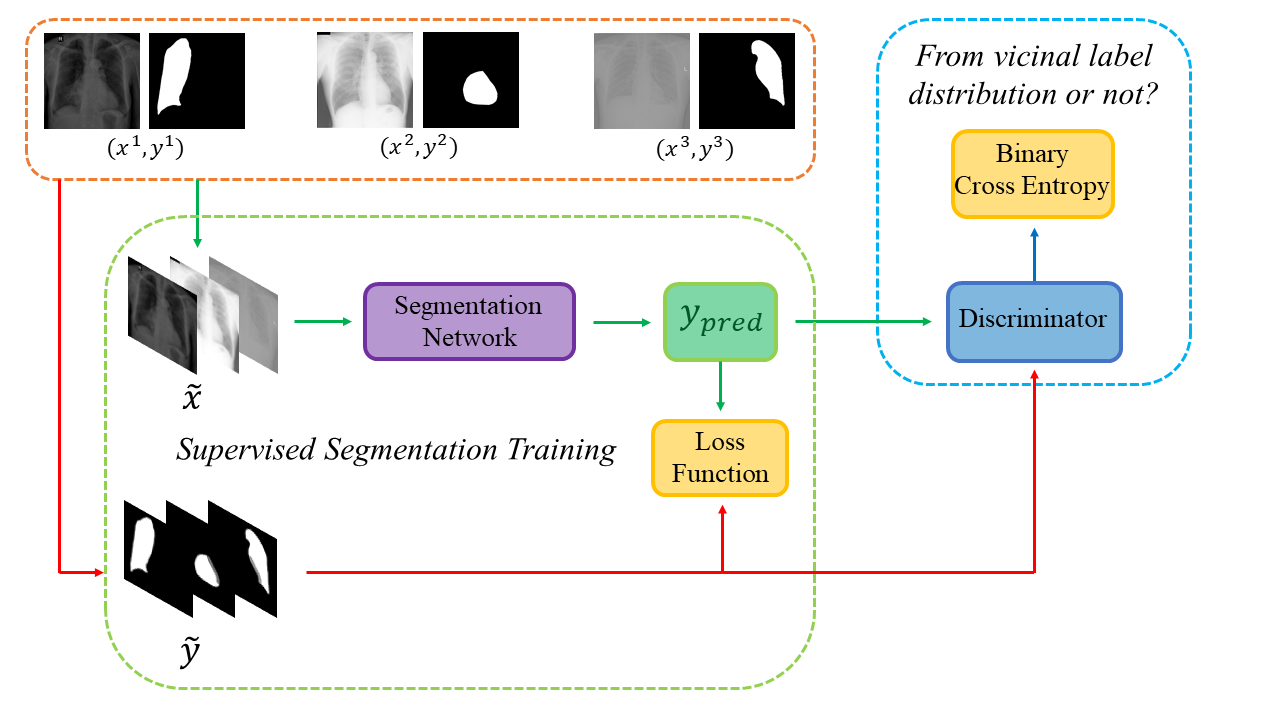}
    \caption{Illustration of adversarial training pipeline. $(\tilde{x}, \tilde{y})$ is generated by the  Eq.~\ref{eq:vic_x} and Eq.~\ref{eq:vic_y}. Same as Fig.~\ref{fig:train}, the background mask is not shown in the figure and we use grayscale images to visualize the vicinal labels. The segmentation network is trained with $(\tilde{x}, \tilde{y})$ in a supervised fashion. $y_{pred}$ is the output of the segmentation network, which is the concatenation of $(K+1)$ probability maps. An auxiliary discriminator is trained to identify whether $y_{pred}$ is sampled from the vicinal distribution, i.e.~discriminate $y_{pred}$ against $\tilde{y}$. The segmentation network and the discriminator are trained alternatively. See Eq.~\ref{eq:seg} and Eq.~\ref{eq:dis} for details.}
    \label{fig:adv}
\end{figure*}

\subsubsection{Extension to Adversarial Training}
\label{sec:vrm_adv}
Compared with previous works in PSL \cite{gonzalez2018multi,petit2018handling,zhou2019prior,dmitriev2019learning,fang2020multi,shi2021marginal}, VLUU can be potentially further improved through adversarial training. Adversarial training was first proposed by \cite{goodfellow2014generative} and several breakthroughs have been made through adversarial training in medical image segmentation \cite{dai2018scan,moeskops2017adversarial,chen2018adversarial,han2018spine}. However, adversarial training for semantic segmentation is ill-defined when the ground truth labels are missing \cite{luc2016semantic}. As VLUU can transform the partially supervised problem into a fully supervised one, it is natural to consider incorporating VLUU and adversarial training. Note, having complete labels during training gives VLUU unparalleled advantages in utilizing some well-known properties of adversarial training, which is difficult for most partially supervised methods. 

In standard adversarial training, the segmentation network and the discriminator play a zero-sum game. The discriminator is trained to discriminate the prediction masks produced by the segmentation network from the ground truth masks. Meanwhile, the segmentation network is trained to confuse the discriminator by producing realistic prediction masks. Adversarial training benefits from the human structure similarity as it makes the unknown true label distributions easier to be caught by the discriminator than for general objects \cite{dong2019neural}. In other words, there is smaller instance-wise variation in the size, shape, and location of human organs (or structures), as shown in Fig.~\ref{fig:1}, than for general objects.

Assume the segmentation network is parameterized by $f_{\theta}$ and the discriminator is parameterized by $g_{\phi}$. Given $\phi$ fixed, $\theta$ is updated by minimizing
\begin{equation}
    \mathcal{L}_\theta = \mathcal{L}_{seg}(f_{\theta}(\tilde{x}), \tilde{y}) - \lambda \log g_{\phi}(f_{\theta}(\tilde{x})),
\label{eq:seg}
\end{equation}
where $\mathcal{L}_{seg}$ is the multi-class cross-entropy loss for standard supervised semantic segmentation and $\lambda$ controls the weight of the adversarial loss. Given $\theta$ fixed, $\phi$ is updated by minimizing
\begin{equation}
    \mathcal{L}_\phi = -\log g_{\phi}(\tilde{y}) - \log(1 - g_{\phi}(f_{\theta}(\tilde{x})).
\label{eq:dis}
\end{equation}
See Fig.~\ref{fig:adv} for the illustration of adversarial training with the vicinal examples. We denote VLUU with adversarial training as VLUU-ADV.

Further, \emph{continuous} vicinal labels have a built-in advantage in stabilizing adversarial training. They alleviate the problem that there commonly is a clear discrepancy between the \emph{discrete} distribution of the ground truth and the \emph{continuous} distribution of the pixel-wise predictions, which can be easily caught by the discriminator \cite{luc2016semantic} and destabilize training, leading to oscillating parameters \cite{salimans2016improved}. Last but not least, with adversarial training, VLUU can further utilize unlabeled data in addition to the partially labeled data. For the interested readers, the problem formulation and application of adversarial training for SSL can be found in \cite{dong2018unsupervised}. 

\section{Theoretical Analysis}
\label{sec:ana}
In this section, we will discuss the theoretical advantages and limitations of the proposed framework.  

\subsection{Enlarged Sample Space}
\label{sec:ana_sample}
One of the main challenges for DL is overfitting caused by data scarcity. In this work, there are two aspects of data scarcity: 1) each image has an incomplete label, and 2) each $S_i$ has only a small number of images. For 1), Eq.~(\ref{eq:vic_y}) and Eq.~(\ref{eq:vic_x}) generate fully labeled vicinal example pairs, thus traditional end-to-end training techniques in supervised learning can finally be applied. 

For 2), with limited training data, state-of-the-art CNN architectures can easily overfit to the training data. Let us first isolate the randomness effect caused by the Dirichlet distribution by setting $w_i = \frac{1}{K}$. The proposed framework enlarges the sample space from $\sum_i n_i$ partially labeled examples to $\prod_i n_i$ fully labeled example pairs. In fact, given $\{(x_{i}, y_{i})\}_{i=1}^K$, $\text{Dirichlet}(\bm{\alpha})$ can theoretically generate an infinite number of $\tilde{y}$ determined by $\bm{w}$. We efficiently mitigate the overfitting problem by enlarging the sample space of $\tilde{S}$.

\subsection{Label Smoothing}
\label{sec:ana_lbl}
In semantic segmentation tasks, labels usually follow a discrete distribution, while Eq.~(\ref{eq:vic_y}) defines a continuous distribution. Even though the application of continuous label distributions is rare in semantic segmentation, they have led to recent breakthroughs in image classification \cite{szegedy2016rethinking,zhang2018mixup}. We expect Eq.~(\ref{eq:vic_y}) can improve the robustness of the model as suggested by recent theoretical analysis of continuous label distributions \cite{muller2019does}. 

\subsection{Computational Cost}
\label{sec:ana_cost}
The training process of the proposed framework is almost identical to the training process for a fully supervised task, i.e. given a segmentation network, there is no additional optimization cost such as multi-stage training \cite{zhou2019prior}. Similarly, the proposed method utilizes the same memory footprint in terms of CNN weights. As a comparison, a semi-supervised method such as label propagation and knowledge transfer will require the training of multiple segmentation networks to generate pseudo-labels. For the proposed method, the major overheads arising from the data generation process are the random sampling and the element-wise operations on low-dimensional arrays, which are negligible compared to the backpropagation cost. Eq.~(\ref{eq:vic_y}) and Eq.~(\ref{eq:vic_x}) can be easily implemented by any scientific computing frameworks supporting broadcasting, such as NumPy, PyTorch, and TensorFlow.

\subsection{Limitations}
The main purpose of the proposed framework is to train DL-based segmentation models with partial labels in an efficient way. As discussed in Sec.~\ref{sec:vrm_def}, the design of Eq.~(\ref{eq:vic_y}) and Eq.~(\ref{eq:vic_x}) makes a strong assumption that all classes of interest are present in each image and there is no hierarchical relationship between the semantic classes, i.e. the classes of interest are mutually exclusive, e.g.~organs in the same body part or sub-structures under the same structure. The situation where the semantic classes have a hierarchical structure, e.g.~liver and liver tumor, is beyond the scope of discussion.

Note, the proposed framework is designed for DL tasks on only a few images without complete annotations. When fully labeled data is available, state-of-the-art supervised and semi-supervised methods have obvious advantages over the proposed framework. However, the proposed framework fills the gap when supervised and semi-supervised methods fail. 

\section{Empirical Analysis}
\label{sec:exp}
The purposes of the experimental design are threefold. First, there is no known empirical study of PSL with limited data. We want to investigate the impact of limited partial labels on DL. Second, we want to systematically evaluate the robustness of the representative partially supervised methods in a controlled environment. Third, we want to demonstrate the effectiveness of VLUU in situations where only a few partially labeled images are available. Thus, the choice of the network backbone or loss function is \textit{independent} of the proposed learning framework. In addition, the simulated experiments are solely to demonstrate the challenges of data scarcity in a controllable environment. We consider two medical image segmentation tasks, chest organ segmentation and optic disc-and-cup segmentation.

\subsection{Chest Organ Segmentation}
\label{sec:exp_cos}
The task of chest organ segmentation is a simple benchmark task in medical image segmentation. In this task, we consider three semantic classes, namely \textit{left lung}, \textit{right lung}, and \textit{heart}. We can easily control the environment to get an insight into the impact of the limited partial labels on various representative partially supervised methods and the efficiency of VLUU. Without specification, the experimental comparison is conducted in such a way that different models use the same network backbone, loss function, training strategy, and the set of hyperparameters. 

\subsubsection{Datasets}
We use two public datasets to simulate the realistic situations that each partially labeled dataset is annotated for a different semantic class and is collected from an independent source. Unlike \cite{shi2021marginal}, which only consider partially labeled datasets, we use two fully labeled datasets to better understand the influence of partial labels.

The \textbf{JSRT} dataset, released by the Japanese Society of Radiological Technology (JSRT), is a benchmark dataset for chest organ segmentation \cite{jsrt}. JSRT contains 247 grayscale CXRs with pixel-wise annotations of lungs and hearts. Each CXR has a size of $2048 \times 2048$.

The \textbf{Wingspan} dataset was collected by Wingspan Technology for the study of transfer learning and unsupervised domain adaptation in chest organ segmentation \cite{dong2018unsupervised}. Wingspan contains 221 grayscale CXRs with pixel-wise annotations of lungs and hearts. The CXRs were collected from 6 hospitals with different imaging protocols. Wingspan expresses a large variety in the data modalities including brightness, contrast, position, and size.

\begin{figure*}[t]
    \centering{
    \begin{subfigure}[t]{0.16\textwidth}
        \centering
        \includegraphics[width=\textwidth]{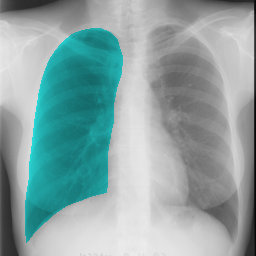}
        \caption{\texttt{L}}
    \end{subfigure}
    \begin{subfigure}[t]{0.16\textwidth}
        \centering
        \includegraphics[width=\textwidth]{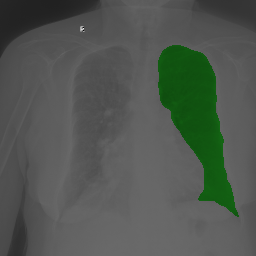}
        \caption{\texttt{R}}
    \end{subfigure}
    \begin{subfigure}[t]{0.16\textwidth}
        \centering
        \includegraphics[width=\textwidth]{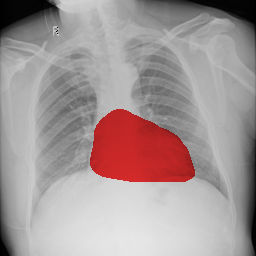}
        \caption{\texttt{H}}
    \end{subfigure}
    \begin{subfigure}[t]{0.16\textwidth}
        \centering
        \includegraphics[width=\textwidth]{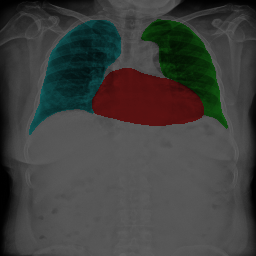}
        \caption{\texttt{T}}
    \end{subfigure}
    \caption{Visual comparison of three partially labeled training sets and one fully labeled test set with corresponding annotations.}
    \label{fig:data_vis}
    }
\end{figure*}

We use three partially labeled datasets as the training set and one fully labeled as the test set, where the four datasets are collected from four different sources. We choose this setup to simulate the practical scenarios where dataset shift exists, which is a challenging situation for DL models. We use the JSRT dataset as the left lung dataset, denoted as \texttt{L}. We use a subset of the Wingspan dataset containing 18 CXRs as the right lung dataset, denoted as \texttt{R}. We use another subset of the Wingspan dataset containing 18 CXRs as the right lung dataset, denoted as \texttt{H}. We use the rest of the Wingspan dataset as the fully labeled test set, which contains 185 CXRs, and denote it as \texttt{T}. The visual comparison of the data modalities of the four sets can be viewed in Fig.~\ref{fig:data_vis}. Note, all four sets are collected from 4 different sources (hospitals with different imaging protocols).

\subsubsection{Baseline Models}
For a fair comparison, we use the same segmentation network for all methods, which is a FCN \cite{long2015fully} with a ResNet18 \cite{he2016deep} backbone. Considering the data scarcity situation, we choose ResNet-FCN as it can both achieve promising results on chest organ segmentation tasks \cite{dong2018unsupervised} and avoid overfitting. We choose the following representative approaches as the baseline models.

\textbf{Fully Supervised Learning Approach} To illustrate the effect of limited partial labels on DL models, we consider two practical approaches in computer vision that are commonly used during large-scale training. As discussed in Sec.~\ref{sec:related_psl}, two methods can be used to train end-to-end methods in a supervised fashion. The first one is to categorize the uncertain (missing) classes as the background in the training, which can be considered as a naive solution with \textit{noisy labels}. We denote the first baseline as MBG because we mix uncertain pixels with the background pixels. The second baseline is to ignore the cross-entropy of the missing classes during the backpropagation. This method is motivated by the nature of multi-task learning for neural networks. We denote this method as IMBP. It is worth mentioning that MBG and IMBP further motivate many recently proposed methods for PSL \cite{gonzalez2018multi,petit2018handling,fang2020multi}.

\textbf{Semi-Supervised Learning Approach} We adopt a strong SSL baseline, label propagation (LP) \cite{iscen2019label}, to solve PSL problem. LP is not an end-to-end method as there are multiple training stages. It first generates \textit{noisy} pseudo-labels for the unlabeled classes based on the partially labeled data. Then the pseudo-labels and ground truth labels are trained together to make the final prediction. However, the quality of the noisy pseudo-labels is highly dependent on the quality of the partially labeled examples and noisy labels might harm the later fine-tuning stage. In this work, we use $K$ independent binary segmentation networks to generate the initial pseudo-labels. 

\textbf{Multi-Task Learning Approach} A classical way to address MTL problems is to fuse knowledge extracted from each individual sub-task~\cite{vandenhende2021multi}, which is also known as knowledge transfer (KT) in the \textit{transfer learning} literature. We train $K$ binary segmentation networks with a shared ResNet feature extractor but independent deconvolutional layers. We alternatively optimize $K$ binary segmentation networks on the corresponding $K$ partially labeled datasets. The final prediction masks is generated by fusing $K$ binary prediction masks. For each pixel, if all classes of interest have probabilities less than the threshold 0.5, we treat it as the background. Otherwise, the pixel is categorized as the class with the highest probability.

\textbf{Partially Supervised Learning Approach} We consider the state-of-the-art partially supervised method \textit{exclusion loss} (EL) \cite{shi2021marginal}, which is designed for the same problem formulation in Sec.~\ref{sec:vrm_def}. EL has shown superior performance over recent partially supervised methods, such as PaNN \cite{zhou2019prior} and PIPO-FAN \cite{fang2020multi}, in all aspects. Unlike EL, recent partially supervised methods rely on either large training data \cite{gonzalez2018multi,petit2018handling,dmitriev2019learning,fang2020multi} or fully labeled data as a prior \cite{zhou2019prior}, which are not applicable for some situations. Similar to our approach, EL can be applied to any existing segmentation networks. So they can be compared with VLUU in a fair setting.

\subsubsection{Implementation}
\label{sec:exp_cos_imp}
The image size is fixed to be $256 \times 256$. We pre-process the raw images by instance normalization. Given an image $x$, we obtain the normalized image $\hat{x}$ by
$\hat{x}^{ij} =  \frac{x^{ij} - \mu(x)}{\sigma(x)}$,
where $(i,j)$ is the position of the pixel in a $256 \times 256$ image, and $\mu$ and $\sigma$ are the mean and standard deviation of the pixels of $x$. In this study, we do not apply other pre-processing techniques as there is no obvious difference in the relative position of objects in each image and the proposed framework is robust against slight misalignment. In practice, when partially labeled datasets are acquired from different imaging protocols, pre-processing techniques such as registration, resizing, and cropping are necessary. There are no fully labeled images in the training set and we consider the setting where each training image only has an annotation of one semantic class, as described in Sec.~\ref{sec:vrm_def}.

All experiments are implemented in PyTorch on an NVIDIA Tesla V100. For a fair comparison, all the networks are initialized with the \textbf{same} random seed and trained from scratch. We use a standard multi-class cross-entropy as the loss function for all the experiments. The batch size is 8. The models are trained to converge with an Adam \cite{kingma2015adam} optimizer and a fixed learning rate of $10^{-3}$. The performance metric in this study is the mean Intersection-Over-Union (mIOU) between the prediction masks and ground truth masks over the three classes of interest. For VLUU, we set $\alpha = 0.1$.

\subsubsection{Comparison Under Small-Scale Data}
\label{sec:exp_res}
\begin{figure*}[t]
\centering
\begin{subfigure}[t]{0.16\textwidth}
    \includegraphics[width=\textwidth]{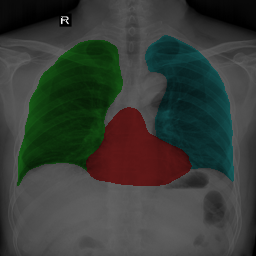}
    \includegraphics[width=\textwidth]{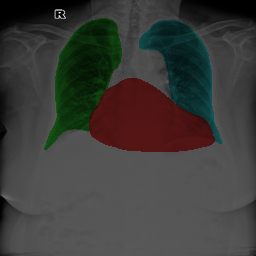}
    \includegraphics[width=\textwidth]{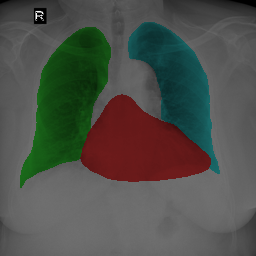}
    \caption{GT}
\end{subfigure}
\rulesep
\begin{subfigure}[t]{0.16\textwidth}
    \includegraphics[width=\textwidth]{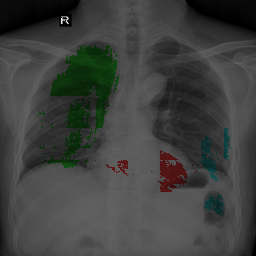}
    \includegraphics[width=\textwidth]{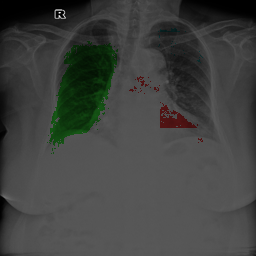}
    \includegraphics[width=\textwidth]{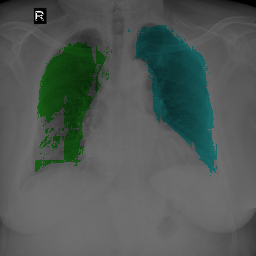}
    \caption{MBG}
\end{subfigure}
\begin{subfigure}[t]{0.16\textwidth}
    \includegraphics[width=\textwidth]{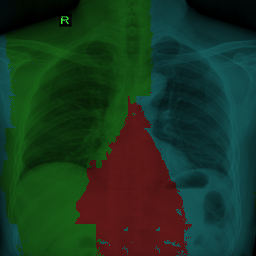}
    \includegraphics[width=\textwidth]{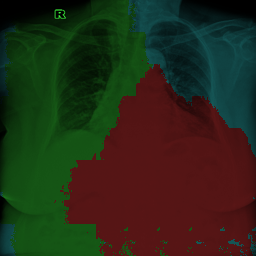}
    \includegraphics[width=\textwidth]{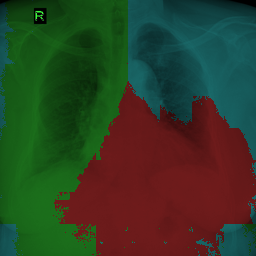}
    \caption{IMBP}
\end{subfigure}
\rulesep
\begin{subfigure}[t]{0.16\textwidth}
    \includegraphics[width=\textwidth]{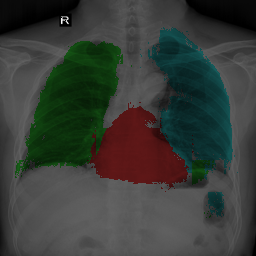}
    \includegraphics[width=\textwidth]{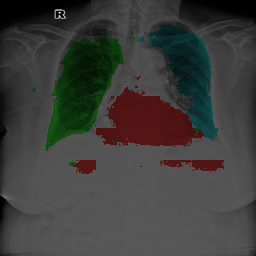}
    \includegraphics[width=\textwidth]{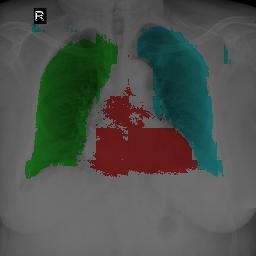}
    \caption{EL}
\end{subfigure}
\begin{subfigure}[t]{0.16\textwidth}
    \includegraphics[width=\textwidth]{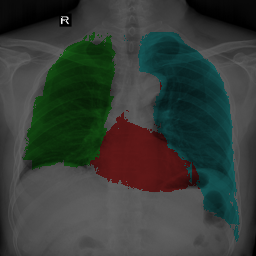}
    \includegraphics[width=\textwidth]{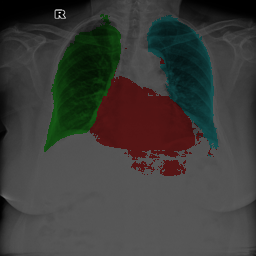}
    \includegraphics[width=\textwidth]{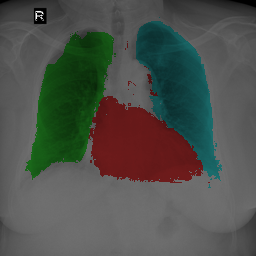}
    \caption{VLUU}
\end{subfigure}
\caption{Qualitative comparison of end-to-end methods on partially supervised chest organ segmentation with $n=15$. GT denotes the ground truth. The segmentation network is ResNet-FCN. $n$ denotes the number of images in each partially labeled dataset. Traditional training strategies for supervised learning, such as (b) MBG and (c) IMBP, fail for PSL. Compared with (d) EL, (e) VLUU generates more realistic organ masks.}
\label{fig:vis_cxr}
\end{figure*}

Because the partially labeled datasets are collected from different sources, we will focus on the challenges of data scarcity and class imbalance. As we want to examine how the size of the partially labeled datasets affects the DL models, we only include $n$ examples of each partially labeled dataset for a quantitative comparison. We provide the performance of the segmentation networks trained on the same training data but with complete annotations as an \textit{Oracle} to provide a reference for the performance. The results are shown in Table~\ref{tab:scarcity}. Supervised methods fail to address the partial labels due to overfitting. As shown in Fig.~\ref{fig:vis_cxr}, MBG tends to predict every pixel as the background while IMBP fails to identify the background, which follows the discussion in Sec.~\ref{sec:related_psl}. 
LP, KT, and EL mitigate the partial labels problem from different perspectives and achieve much better performance than supervised methods. However, these seminal methods suffer from the limited training data and multi-source domain shift.
Among the baseline methods, LP is the most computationally expensive method as it requires considerably more training time and memory footprint than all other methods. In addition, LP is more sensitive to the size of the training set. In practice, semi-supervised models expect a large set of unlabeled data, which is not aligned with the problem formulation in this work. 
Compared with semi-supervised methods, MTL methods usually consume a much smaller memory footprint depending on the number of shared layers. By comparing KT and VLUU, we can see that VLUU has more shared neural architectures than KT, which can reduce the memory footprint and substantially improve the model performance. 
As the state-of-the-art partially supervised method, EL purely relies on using a modified loss function to extract knowledge from the training. When there is not enough training data, EL performs worse than KT and VLUU.
In contrast to the baseline methods, VLUU achieves the best performance on small-scale data. Without acquiring any new supervision, VLUU incorporating a coarse anatomical knowledge by uniquely utilizing human structure similarity.

\begin{table*}[t]
\centering{
\setlength{\tabcolsep}{1em}
\footnotesize{
\begin{tabular}{lcccc}
\hline
Method & Type & $n=5$ & $n=10$ & $n=15$\\ \Xhline{4\arrayrulewidth}
MBG & SL & 0.3187 & 0.3221 & 0.2715 \\ \hline
IMBP~\cite{gonzalez2018multi} & SL & 0.2715 & 0.3161 & 0.3218\\ \hline
LP~\cite{iscen2019label} & SSL & 0.5821 & 0.7444 & 0.7588 \\ \hline
KT~\cite{vandenhende2021multi} & MTL & 0.6478 & 0.6686 & 0.7071 \\ \hline
EL~\cite{shi2021marginal} & PSL & 0.6306 & 0.6591 & 0.7506 \\ \hline
VLUU & PSL & \textbf{0.7063} & \textbf{0.7462} & \textbf{0.7615} \\ \Xhline{2\arrayrulewidth}
\textit{Oracle} & SL & 0.7860 & 0.8395 & 0.8487 \\ \hline
\end{tabular}
}
}
\caption{Quantitative comparison (mIOU) on partially supervised chest organ segmentation with small-scale data. The segmentation network is ResNet-FCN. $n$ denotes the number of images in each partially labeled dataset.}
\label{tab:scarcity}
\end{table*}

It is worth mentioning that, MBG, IMBP, EL, and VLUU are end-to-end methods, i.e.~they do not require any auxiliary NNs or multi-stage training procedures. We provide the qualitative comparison of end-to-end methods in Fig.~\ref{fig:vis_cxr}. VLUU tends to output more realistic masks than the STOA method EL in terms of the location and shape.

\subsubsection{Comparison Under Class Imbalance}
\label{sec:exp_imbalance}
Considering the availability of the medical data and the difficulty of annotating certain organs or structures, we simulate the class imbalance situations in PSL. Here, we use $\eta$ to control the class imbalance. As the heart is more difficult to annotate than the two lungs \cite{dai2018scan}, we set the partially labeled dataset for the heart (\texttt{H}) to have $n=5$ and the partially labeled datasets for the two lungs (\texttt{L} and \texttt{R}) to both have $\eta n$ examples. The results are shown in Table~\ref{tab:imbalance}. Compared with Table~\ref{tab:scarcity}, the class imbalance does have a severe negative impact on the baseline methods MBG, IMBP, and KT, as more training data could even decrease the performance. While LP, EL, and VLUU could benefit from more training data, LP achieves much lower performance than EL and VLUU. VLUU can generally achieve comparable performance with EL while outperforming EL by a large margin with small $n$. Compared with the baseline methods, VLUU mitigates the class imbalance by utilizing human structure similarity to generate a balanced vicinal label distribution. 

\begin{table*}[t]
\centering{
\setlength{\tabcolsep}{1em}
\footnotesize{
\begin{tabular}{lcccc}
\hline
Method & Type & $\eta=1$ & $\eta=2$ & $\eta=3$\\ \Xhline{4\arrayrulewidth}
MBG & SL & 0.3187 & 0.3633 & 0.3433 \\ \hline
IMBP~\cite{gonzalez2018multi} & SL & 0.2715 & 0.3052 & 0.3029 \\ \hline
LP~\cite{iscen2019label} & SSL & 0.5821 & 0.6344 & 0.6555 \\ \hline
KT~\cite{vandenhende2021multi} & MTL & 0.6478 & 0.6511 & 0.6446 \\ \hline
EL~\cite{shi2021marginal} & PSL & 0.6306 & 0.7263 & 0.7347 \\ \hline
VLUU & PSL & \textbf{0.7063} & \textbf{0.7268} & \textbf{0.7365} \\ \Xhline{2\arrayrulewidth}
\textit{Oracle} & SL & 0.7860 & 0.8208 & 0.8340 \\ \hline
\end{tabular}
}
}
\caption{Quantitative comparison (mIOU) of methods on chest organ segmentation with class imbalance. The segmentation network is ResNet-FCN. $\eta$ denotes the ratio of the number of images in the dataset \texttt{L} or \texttt{R} to the number of images in the dataset \texttt{H}.}
\label{tab:imbalance}
\end{table*}

\subsubsection{Ablation Studies}
\textbf{Impact of Network Complexity}
Under the data scarcity challenge, the complexity of the segmentation network will usually play an important role. The network complexity is determined by the number of parameters and the network architecture. For supervised tasks, U-Net should outperform ResNet-FCN because U-Net has more parameters than ResNet-FCN\footnote{U-Net has 38.8M parameters and FCN with a ResNet18 backbone has 13.3M parameters.} and a better network architecture design for medical image segmentation tasks \cite{ronneberger2015u}. Clearly, there is a trade-off in the network selection between the network complexity and network performance when the partially labeled datasets are small. Here, we evaluate VLUU with both FCN and U-Net, and results are shown in Table~\ref{tab:complexity}. We hypothesize that complex networks have a negative impact on VLUU when there is only limited data. Given a small amount of training data, complex networks could have both performance gain due to more parameters and delicate architectures, and performance drop due to overfitting, depending on the amount of training data.

\begin{table*}[t]
\centering{
\setlength{\tabcolsep}{1em}
\footnotesize{
\begin{tabular}{lccc}
\hline
Network & $n=5$ & $n=10$ & $n=15$\\ \Xhline{4\arrayrulewidth}
FCN~\cite{long2015fully} & \textbf{0.7063} & \textbf{0.7462} & 0.7615 \\ \hline
U-Net~\cite{ronneberger2015u} & 0.5411 & 0.7261 & \textbf{0.7799} \\ \hline
\end{tabular}
}
}
\caption{The impact of network complexity on VLUU with ResNet-FCN as the segmentation network. $n$ denotes the number of images in each partially labeled dataset.}
\label{tab:complexity}
\end{table*}

\begin{figure}[t]
    \centering
    \includegraphics[width=0.6\textwidth]{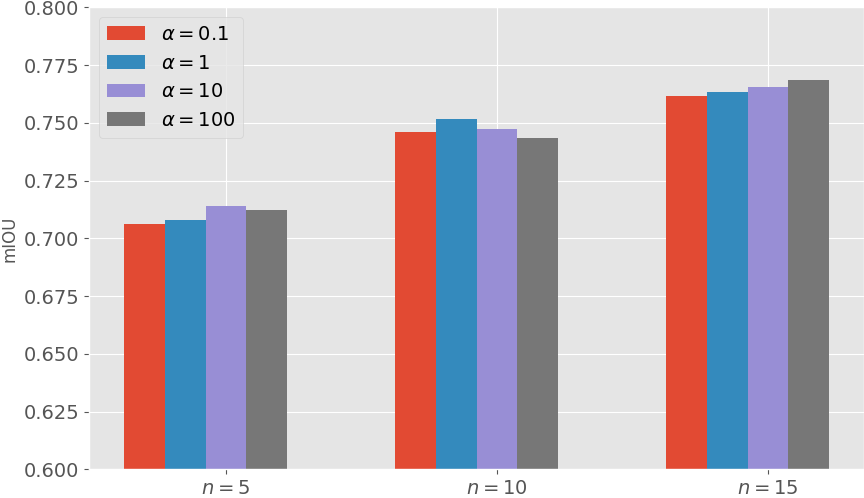}
    \caption{Sensitivity of $\alpha$ to $n$. Overall, VLUU is robust for various $\alpha$.}
    \label{fig:alpha}
\end{figure}

\textbf{Sensitivity to $\bm{\alpha}$}
The performance of a ResNet-FCN trained by VLUU with different $\alpha$ is shown in Fig.~\ref{fig:alpha}. Overall, VLUU is not sensitive to $\alpha$ as there are only small differences between the performance for different $\alpha$ values. Note, $\text{Dirichlet}(\bm{\alpha})$ is asymptotically close to a uniform distribution when $\alpha \to \infty$, i.e. $w_i = \frac{1}{K}$. In addition, there is a trade-off in selecting the optimal $\alpha$. Small $\alpha$ indicates a larger variation in the label distribution, which means larger uncertainty. So, for tasks such as chest organ segmentation where the organs have relatively fixed locations and similar shapes, a large $\alpha$ might help. However, a small $\alpha$ should be more robust as it introduces more uncertainty when $K$ is large. In this work, we use $\alpha = 0.1$ for consistency.

\textbf{Effect of Random Initiation} To examine the sensitivity of the proposed framework to the effect of random initiation, we repeat the experiments in Table~\ref{tab:scarcity} for EL and VLUU for 5 times each. This time, the backbone network is randomly initiated at each time. Unlike the results in Table~\ref{tab:scarcity}, which are the highest mIOU, we report the mean and standard deviation of mIOUs in Table~\ref{tab:random}. Compared with the loss-based partially supervised method EL, the label-based partially supervised method VLUU is more robust with smaller standard deviation.

\begin{table*}[t]
\centering{
\setlength{\tabcolsep}{1em}
\footnotesize{
\begin{tabular}{lccc}
\hline
Method & $n=5$ & $n=10$ & $n=15$ \\ \Xhline{4\arrayrulewidth}
EL~\cite{shi2021marginal} & 0.6313 $\pm$ 0.1997 & 0.2587 $\pm$ 0.3966 & 0.7506 $\pm$ 0.1576 \\ \hline
VLUU & 0.7058 $\pm$ 0.1226 & 0.7399 $\pm$ 0.1200 & 0.7609 $\pm$ 0.1036 \\\hline
\end{tabular}
}
}
\caption{Robustness of VLUU under different random initiations. The performance (mean mIOU $\pm$ standard deviation) of VLUU is more stable than the performance of EL.}
\label{tab:random}
\end{table*}

\textbf{Adversarial Training}
For VLUU-ADV, we use a standard ResNet binary classifier as the discriminator as we use a ResNet-FCN as the segmentation network. In fact, the choice of the discriminator is a research question in its own right \cite{dong2019neural}. \cite{motiian2017few} shows that having the same backbones for the segmentation network and the discriminator can increase the stability of adversarial training. We follow the training scheme in Sec.~\ref{sec:vrm_adv}, where the adversarial loss \cite{luc2016semantic} in Eq.~(\ref{eq:seg}) is weighted by $\lambda = 0.001$. We report the results of VLUU and VLUU-ADV in Table~\ref{tab:adv}, where VLUU-ADV shows slightly better results than VLUU. We conclude that ADV can be used as an add-on module for VLUU with \textit{appropriate} $\alpha$ and \textit{delicate} design of the network architecture for the discriminator. 

\begin{table*}[t]
    \centering{
    \setlength{\tabcolsep}{0.8em}
    \footnotesize{
    \begin{tabular}{lccc}
    \hline
    Method & $n=5$ & $n=10$ & $n=15$\\ \Xhline{4\arrayrulewidth}
    VLUU & 0.7063 & \textbf{0.7462} & 0.7615 \\ \hline
    VLUU-ADV & \textbf{0.7171} & 0.7412 & \textbf{0.7630} \\ \hline
    \end{tabular}
    }
    }
    \caption{Quantitative comparison (mIOU) between VLUU and VLUU-ADV with ResNet-FCN as the segmentation network. $n$ denote the number of images in each partially labeled dataset.}
    \label{tab:adv}
\end{table*}

\subsection{Optic Disc-and-Cup Segmentation}
\label{sec:exp_op}
In addition to chest organ segmentation, another task where all classes of interests are present in each image is the optic disc-and-cup segmentation. As an important step of early screening of glaucoma, optic disc-and-cup segmentation on the fundus images localizes the optic disc-and-cup for the analysis of the optical nerve head \cite{wang2019ellipse}. An increase in the optic cup-to-disc ratio could be an indicator of the presence of glaucoma \cite{syc2011cup}. The annotation of the optic disc is more difficult than that of the optic cup. In addition, the optic disc and optic cup have a unique geometric property that the optic cup is always enclosed by the optic disc. That is to say, if we want to annotate the optic disc, we have to annotate the optic cup first. Although this is not the standard problem formulation, VLUU can be applied to this situation directly as discussed in Sec.~\ref{sec:vrm_def}.

\subsubsection{Datasets}
\label{sec:exp_op_data}
We use the REFUGE dataset\footnote{https://refuge.grand-challenge.org} to simulate the experiments for optic disc-and-cup segmentation. As there are two classes of interest, there should be at least two partially labeled datasets. However, as explained before, it is less practical to have a partially labeled dataset for optic disc. Instead, we have one larger partially labeled dataset for optic cup (denoted as \texttt{P}) and one smaller fully labeled dataset (denoted as \texttt{F}) as the training set. This motivation behind is twofold. First, the annotation of optic cup requires less human effort and is much cheaper to acquire than the annotation of optic disc. Second, we want to introduce the class imbalance. As REFUGE is collected from multiple sources, we create two sub-datasets from two sources to simulate the dataset shift in the training set. We use the validation set of REFUGE as the test set (denoted as \texttt{T}), which contains 400 fundus images.

\begin{figure*}[th]
    \centering{
    \begin{subfigure}[t]{0.15\textwidth}
        \centering
        \includegraphics[width=\textwidth]{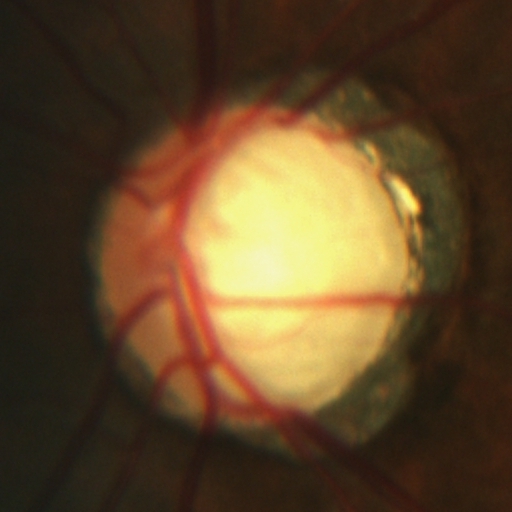}
        \caption{}
    \end{subfigure}
    \begin{subfigure}[t]{0.15\textwidth}
        \centering
        \includegraphics[width=\textwidth]{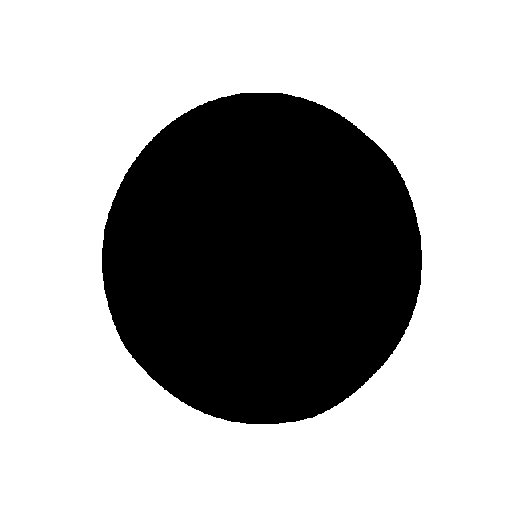}
        \caption{}
    \end{subfigure}
    \begin{subfigure}[t]{0.15\textwidth}
        \centering
        \includegraphics[width=\textwidth]{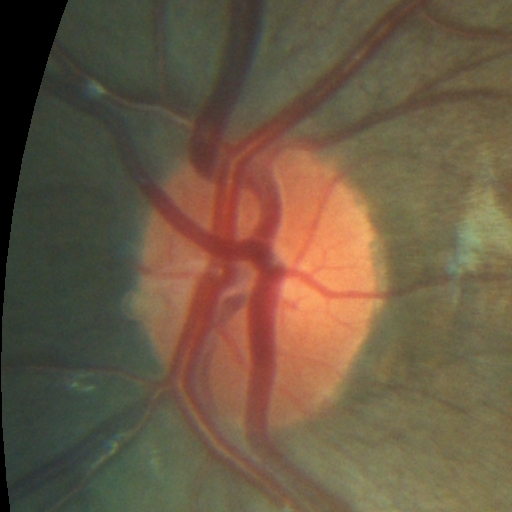}
        \caption{}
    \end{subfigure}
    \begin{subfigure}[t]{0.15\textwidth}
        \centering
        \includegraphics[width=\textwidth]{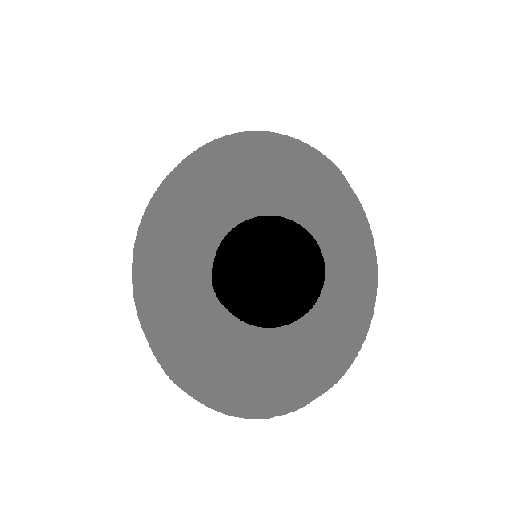}
        \caption{}
    \end{subfigure}
    \begin{subfigure}[t]{0.15\textwidth}
        \centering
        \includegraphics[width=\textwidth]{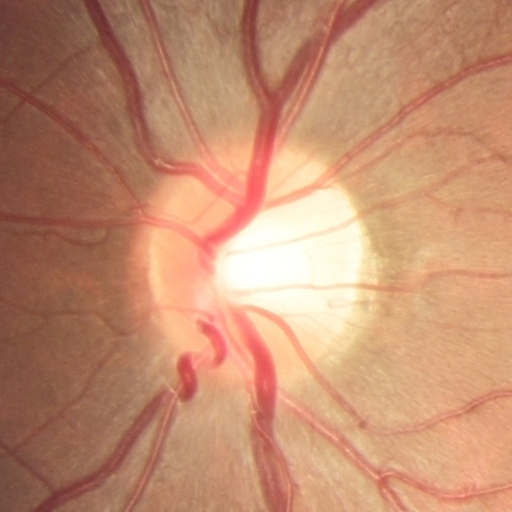}
        \caption{}
    \end{subfigure}
    \begin{subfigure}[t]{0.15\textwidth}
        \centering
        \includegraphics[width=\textwidth]{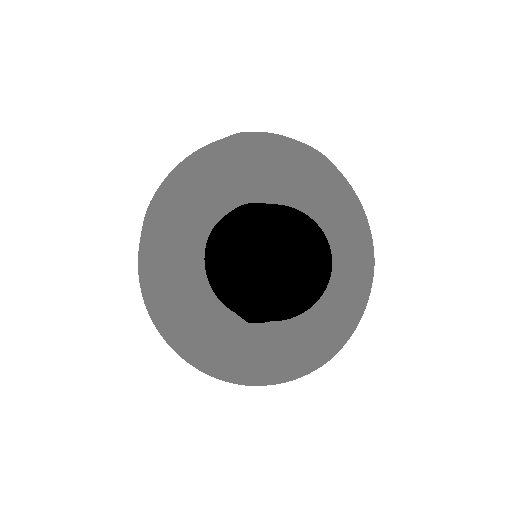}
        \caption{}
    \end{subfigure}
    
    \caption{Visual comparison of the fundus images in the training set and the test set. The training set consists of a partially labeled dataset for optic cup only and a fully labeled dataset for both optic disc and optic cup. (a) A fundus image from the dataset \texttt{P}. (b) The corresponding ground truth mask of (a) with the optic cup annotated as black. (c) A fundus image from the dataset \texttt{F}. (d) The corresponding ground truth mask of (c) with the optic disc annotated as grey and the optic cup annotated as black. (e) A fundus image from the test set \texttt{T}. (f) The corresponding ground truth mask of (e). Note, there are clear dataset shifts among the three datasets.}
    \label{fig:op_data_vis}
    }
\end{figure*}

As REFUGE is collected from multiple sources, the fundus images have various image size. The images are pre-processed by registration, cropping, and resizing to have a fixed resolution of $256 \times 256$. So the pre-processed images contain the whole region of the optical nerve head. See Fig.~\ref{fig:op_data_vis} for examples of the training set and the test set.

\subsubsection{Implementation}
Based on the results in the previous section, we only compare EL and VLUU, as EL and VLUU consistently outperform other methods. In addition, we use a new baseline PaNN \cite{zhou2019prior}. PaNN requires that there is a small fully labeled dataset in the training set to learn the prior, which fits our task setup in Sec.~\ref{sec:exp_op_data} perfectly. Again, for a fair comparison, we use a ResNet-FCN as the network backbone and use the same set of hyperparamters in Sec.~\ref{sec:exp_cos_imp}. The performance metric is the mIOU between the unprocessed\footnote{In practice, the prediction masks could be further improved by image processing techniques.} prediction masks and ground truth masks on optic disc and optic cup.

In contrast to CXRs, the fundus images are color images with RGB channels. To generate a vicinal image, we concatenate two sampled images from the two partially labeled datasets along the RGB channels, i.e.~the vicinal images now have 6 ($3K$ where $K = 2$) channels. Eq.~\ref{eq:vic_x} and Eq.~\ref{eq:vic_y} still hold. In the training of VLUU, we rearrange the training data as two partially labeled datasets. The small fully labeled dataset is split into two sub-datasets containing the same images, where one sub-dataset only contains labels for the optic disc and is treated as the new partially labeled dataest for the optic disc. The other sub-dataset with only labels for the optic cup is added into the partially labeled dataset for the optic cup. 

\begin{table*}[t]
    \centering{
    \setlength{\tabcolsep}{1.2em}
    \footnotesize{
    \begin{tabular}[t]{lcccc}
    \hline
    Method & Type & $n=1$ & $n=2$ & $n=3$ \\ \Xhline{4\arrayrulewidth}
    EL \cite{shi2021marginal} & PSL & 0.1395 & 0.1596 & 0.1991 \\ \hline
    PaNN \cite{zhou2019prior} & PSL / SSL & 0.5976 & 0.5999 & 0.6299 \\ \hline
    VLUU & PSL & \textbf{0.6452} & \textbf{0.7605} & \textbf{0.7945} \\ \Xhline{2\arrayrulewidth}
    \textit{Oracle} & SL & 0.6677 & 0.7045 & 0.7713 \\ \hline
    \end{tabular}
    }
    }
    \caption{Quantitative comparison (mIOU) of PSL methods on partially supervised optic disc-and-cup segmentation with class imbalance. The segmentation network is ResNet-FCN. $n$ denotes the number of images with optic disc annotated.}
    \label{tab:optic}
\end{table*}

\begin{figure*}[th]
\centering
\begin{subfigure}[t]{0.16\textwidth}
    \includegraphics[width=\textwidth]{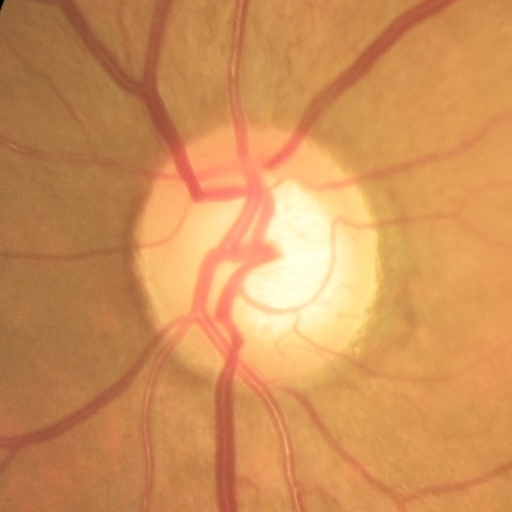}
    \includegraphics[width=\textwidth]{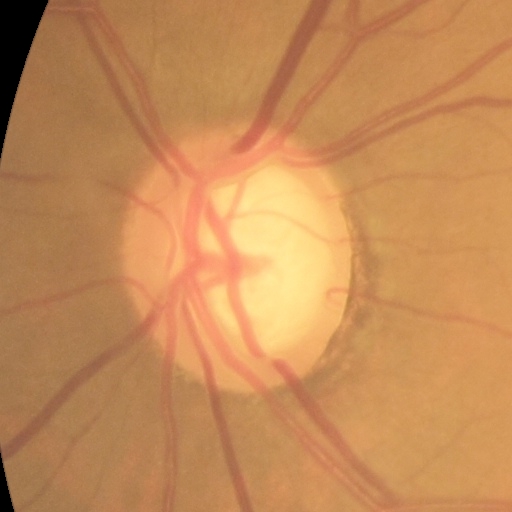}
    \includegraphics[width=\textwidth]{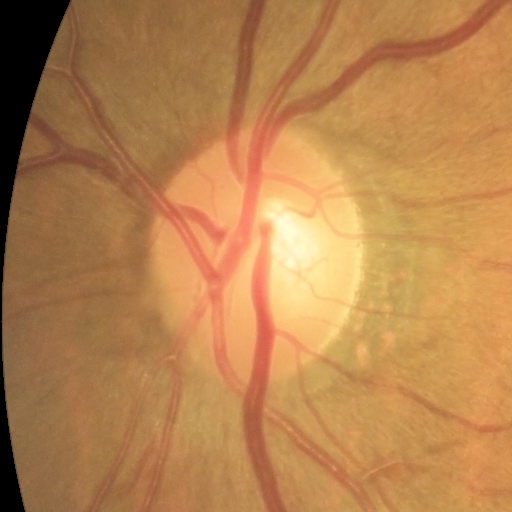}
    \caption{Input}
\end{subfigure}
\begin{subfigure}[t]{0.16\textwidth}
    \includegraphics[width=\textwidth]{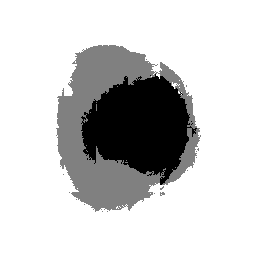}
    \includegraphics[width=\textwidth]{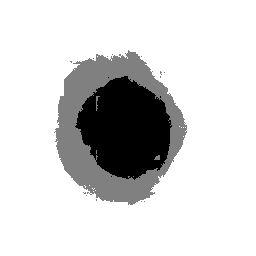}
    \includegraphics[width=\textwidth]{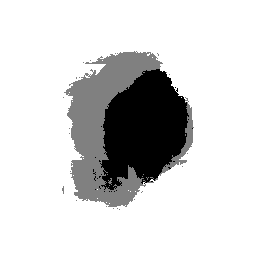}
    \caption{PaNN}
\end{subfigure}
\begin{subfigure}[t]{0.16\textwidth}
    \includegraphics[width=\textwidth]{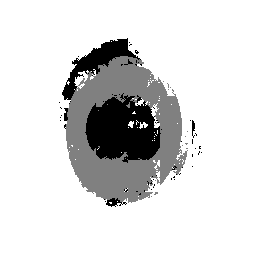}
    \includegraphics[width=\textwidth]{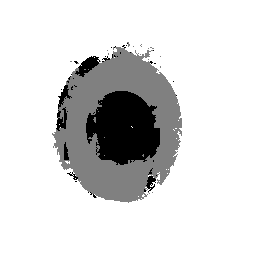}
    \includegraphics[width=\textwidth]{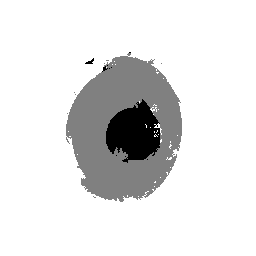}
    \caption{VLUU}
\end{subfigure}
\begin{subfigure}[t]{0.16\textwidth}
    \includegraphics[width=\textwidth]{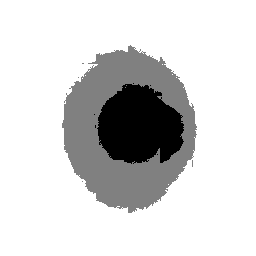}
    \includegraphics[width=\textwidth]{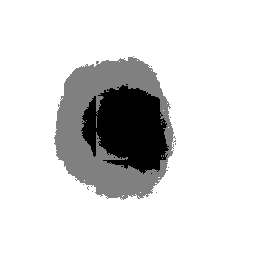}
    \includegraphics[width=\textwidth]{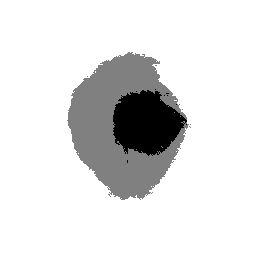}
    \caption{\textit{Oracle}}
\end{subfigure}
\begin{subfigure}[t]{0.16\textwidth}
    \includegraphics[width=\textwidth]{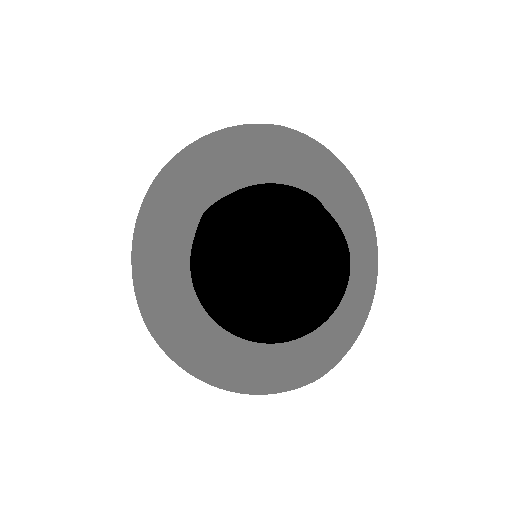}
    \includegraphics[width=\textwidth]{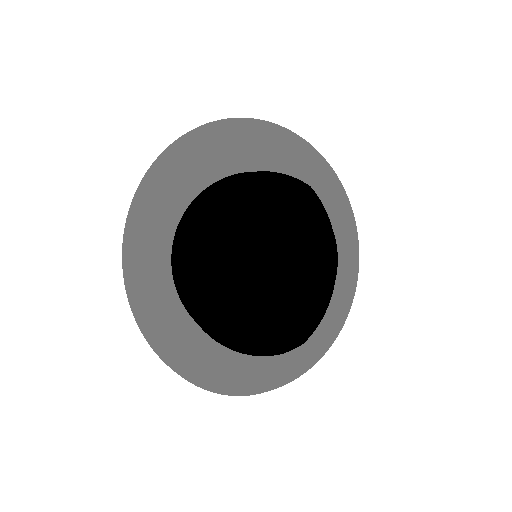}
    \includegraphics[width=\textwidth]{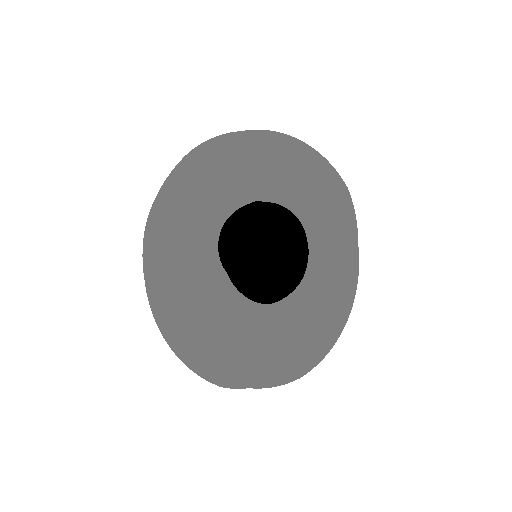}
    \caption{GT}
\end{subfigure}
\caption{Qualitative comparison on partially supervised optic disc-and-cup segmentation with $n=3$. GT denotes the ground truth. The segmentation network is ResNet-FCN. $n$ denotes the number of images with optic disc annotated. A FCN trained with VLUU and partial labels can generate prediction masks which are qualitatively comparable with the masks predicted by a FCN trained with complete labels.}
\label{fig:optic_vis}
\end{figure*}

\subsubsection{Results} 
Compared with the experiments in Sec.~\ref{sec:exp_cos}, we use a more extreme setting to test the limit of partially supervised methods. We use only 10 images from \texttt{P} (i.e.~10 images with optic cup annotated) and $n$ images from \texttt{F} (i.e.~$n$ images with both optic disc and optic cup annotated). There is a severe class imbalance here, as the ratio of the number of labels for cup to the number of labels for disc is $\frac{10 + n}{n}$. The results measured in mIOU between the prediction masks and ground truth masks on optic disc and optic cup are presented in Table~\ref{tab:optic}. With much smaller data size than before, EL fails. Besides, as EL is not designed for fully labeled datasets, the images with complete labels (from \texttt{F}) actually have a negative influence on the training. Meanwhile, PaNN cannot easily learn the image prior based on only a few fully labeled images. VLUU outperforms EL and PaNN by a large margin. Essentially, EL and PaNN do not solve the data scarcity problem, while VLUU can generate new vicinal examples. Moreover, a segmentation network trained with VLUU can even achieve comparable performance with the same network trained with complete labels (i.e.~more supervision). Considering the existence of class imbalance and dataset shift, we conclude that VLUU is more robust on small-scale data. The visual comparison between PaNN, VLUU and \textit{Oracle} is shown in Fig.~\ref{fig:optic_vis}. It can be seen that PaNN generates unrealistic shapes for the optic disc and optic cup if not enough fully labeled data is available learn a reasonable image prior. Note, although VLUU can achieve comparable performance with \textit{Oracle} in numerical results, there are artifacts caused by the uncertainty of the vicinal labels, e.g.~as shown in Fig.~\ref{fig:optic_vis}, VLUU may generate optic cup predictions outside the optic disc. 

\section{Conclusion}
\label{sec:con}
In this paper, we discuss the robustness issue of partially supervised methods under the challenge of data scarcity. We present VLUU, an easy-to-implement framework, for medical image segmentation tasks with only small partially labeled data. Compared with previous methods, VLUU efficiently utilizes the human structure similarity. The experimental results show that VLUU is more robust than state-of-the-art partially supervised methods under various data scarcity situations. Our research suggests a new research direction in label-efficient DL with partial supervision by tackling the problem from the perspective of VRM.

\section*{Acknowledgment}
The authors would also like to thank Huawei, Amazon, and Google for providing cloud computing service for this study. This work was partially funded by the Research Council of Norway grants no. 315029, 309439, and 303514.

\bibliographystyle{elsarticle-num}
\bibliography{refs}
\end{document}